\documentclass{article}

\usepackage[final]{neurips_2022}

\usepackage[utf8]{inputenc} 
\usepackage[T1]{fontenc}    
\usepackage{hyperref}       
\usepackage{url}            
\usepackage{booktabs}       
\usepackage{amsfonts}       
\usepackage{nicefrac}       
\usepackage{microtype}      
\usepackage{xcolor}         
\usepackage{xspace}
\usepackage{subcaption}
\usepackage{amsmath}
\usepackage{makecell}
\usepackage{array}
\usepackage{multirow}
\usepackage{booktabs}
\usepackage[nottoc]{tocbibind}
\usepackage{graphicx}

\def\figref#1{Figure~\ref{fig:#1}}
\def\figlabel#1{\label{fig:#1}\label{p:#1}}

\def\tabref#1{Table~\ref{tab:#1}}

\def\tablabel#1{\label{tab:#1}\label{p:#1}}

\def\secref#1{\S\ref{sec:#1}}
\def\seclabel#1{\label{sec:#1}}

\newcounter{notecounter}
\newcommand{\enotesoff}{\long\gdef\enote##1##2{}}
\newcommand{\enoteson}{\long\gdef\enote##1##2{{
\stepcounter{notecounter}
{\large\bf \hspace{1cm}\arabic{notecounter} $<<<$ ##1: ##2 $>>>$\hspace{1cm}}}}}
\enoteson
\enotesoff

\title{This joke is \texttt{[MASK]}: Recognizing Humor and Offense with Prompting}

\author{
  Junze Li\textsuperscript{†}
  \ \ Mengjie Zhao\textsuperscript{‡}
  \ \ Yubo Xie\textsuperscript{†}
  \ \ Antonis Maronikolakis\textsuperscript{‡}
  \ \  Pearl Pu\textsuperscript{\normalfont{†}}
  \ \  Hinrich Sch\"{u}tze\textsuperscript{‡}\\
  \textsuperscript{†}EPFL
  \ \ \textsuperscript{‡}LMU Munich \\
  \texttt{junze.li@epfl.ch} \ \ \ \texttt{mzhao@cis.lmu.de}
}

\begin{document}
\maketitle
\begin{abstract}

Humor is a magnetic component
in everyday human interactions and communications.
Computationally modeling humor
enables NLP systems to entertain and engage with users.
We investigate the effectiveness
of prompting, a new transfer learning paradigm for NLP,
for humor recognition.
We show that prompting performs similarly to finetuning
when numerous annotations are available, but gives
stellar performance in low-resource humor recognition.
The relationship between humor and
offense is also inspected by applying influence functions
to prompting; we show that models could rely on offense
to determine humor during transfer.
{\textcolor{red}{Disclaimer: This paper contains model outputs that are offensive by
nature.}}
\end{abstract}

\section{Introduction}

Humor is one of the most attractive phenomena in human communication,
providing entertainment and relieving mental stress
\citep{mihalcea2005making,lefcourt2012humor}.
Humor regulates human communication, as a result,
computationally modeling it is expected to improve human-computer
interaction experience \citep{nijholt2003humor,7880297}.

The first attempt to define humor was in ancient Greece, where
ideologists and philosophers considered human laughter during
comedies as a form of scorn \citep{sep-humor}. Superiority theory
of humor considers that the laughter is a
type of superiority over other peoples'
\emph{physical defects or shortcomings}.
Incongruity theory focuses on language
semantics, claiming that incongruous
meanings in the same
context lead to a humor effect
\citep{kant1790critique,schopenhauer1883world}. This
type of humor with the form of ``setup + punchline'' is widely
studied in NLP;
numerous linguistic resources were created
\citep{mihalcea2010computational,bertero2016long,xie2020uncertainty}.

Humor recognition \citep{mihalcea2005making},
striving to identify humorous texts,
is the first step to enable NLP models to understand humor.
\emph{State-of-the-art models} \citep{meaney-etal-2021-semeval}
\emph{rely on transfer learning from
large-scale pretrained language models (PLMs)};
finetuning \citep{devlin2018bert} is a common practice.
In this work, we further investigate
the effectiveness of prompting \citep{GPT3paper,schick20cloze,liu2021pre}
in humor recognition. This is inspired by the fact that
recognize humor of the form  ``setup + punchline''
well matches prompting patterns.
E.g., we can query a language
model to fill in the blank in a pattern
``setup + punchline. \underline{It} \underline{is} \texttt{[MASK]}'',
by answering ``funny'' or ``normal''.
We show that prompting performs comparably to
finetuning when numerous annotations
are available, but
it clearly outperforms finetuning
in low-resource transfer scenarios.

Humor is subjective:
judgment of what is humorous varies across human personalities, cultural
backgrounds, and commonsense knowledge \citep{chen2018humor}.
As in the superiority theory, it is likely that
some humor recognition datasets include
``humorous'' content that is offensive, e.g., offensive to women.
This is undesirable because a machine-learning based NLP system,
e.g., a virtual assistant,
should never
respond to a user query such as ``Tell me a joke!''
with text that is offensive (even if some may view the text
as humorous).
It is thus crucial to
identify, mitigate, and reduce  offense
when modeling humor computationally.

The SemEval-2021 Shared Task
shows that the relationship
between humor and offense is
subtle and challenging to detect \citep{meaney-etal-2021-semeval}.
In this work, we leverage
influence functions (\textsc{If}; \citet{cook1982residuals,koh2017understanding})
to identify offense when building humor recognition systems
based on transfer learning.
For each humorous test example, we locate its influential
training examples and then inspect if they
were offensive.
Since utilizing \textsc{If} does not
incur architectural modifications \citep{koh2017understanding},
we integrate it to both finetuning and prompting.
\textsc{If} identifies humorous and offensive training examples,
in both finetuning and prompting.
However, the accuracy of identification varies
when different resource limitations apply.

We make the following \textbf{contributions}:
\textbf{(i)} We investigate the effectiveness of prompting,
    the new transfer learning paradigm in NLP, for humor recognition.
    Prompting performs similarly to
    finetuning when enough data is available, but gives
    stellar performance in low-resource scenarios.
\textbf{(ii)} We integrate influence functions to finetuning and prompting,
    and show that the identified influential training examples
    indeed contain offensive contents.
    This characterization implies that
    PLM-based humor recognition systems rely on
    offensive contents to determine the presence of humor, which
    is an undesirable behavior and should
    be avoided in computational humor.

\section{Method}
\seclabel{sec:method}
We start with introducing finetuning and prompting for humor recognition.
Next, we introduce procedures of estimating
the influence of individual training examples for a test example.

\textbf{Finetuning}.
We follow the standard practice of finetuning BERT as
\citet{devlin2018bert}. We randomly initialize a linear
classifier layer and then stack it on top of BERT. All the parameters
are then finetuned using the humor recognition dataset.
Vector of \texttt{[CLS]} is used to represent the sentence.

\begin{figure}[t]
    \centering
    \includegraphics[width=0.8\linewidth]{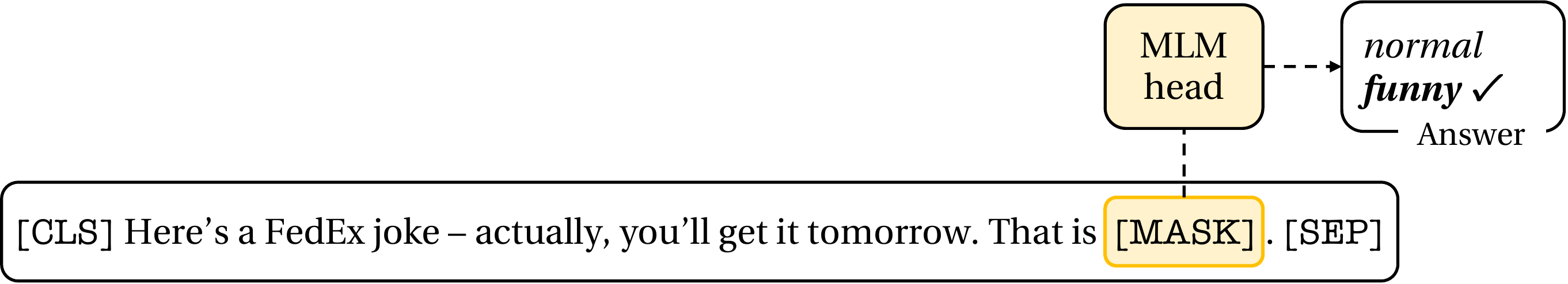}
    \caption{The prompting method for humor recognition.}
    \figlabel{fig:prompting}
\end{figure}

\textbf{Prompting}
is a recent method of utilizing the PLMs \citep{GPT3paper,schick20cloze,liu2021pre}.
Unlike finetuning that randomly initializes a new classifier head, prompting reuses
the masked-language model (MLM) head of BERT, requiring no extra new parameters.
Prompting reformulates a sentence $\mathbf{x}$ using a pattern
$f_{prompt}(\cdot)$ which contains the \texttt{[MASK]} token.
After that, the PLM is asked to fill in the \texttt{[MASK]} token by selecting
a token from $\mathcal{V}$=\{$y_1$, $y_2$\}, using the MLM head. In this work,
we use
$f_{prompt}(\mathbf{x}) =  \mathbf{x} \ \ \underline{That}\ \  \underline{is}\ \  \texttt{[MASK]}\ \underline{.}$
 and $\mathcal{V}$ = \{``funny'',
``normal''\}. \figref{fig:prompting} demonstrates prompting with a
concrete example. To compute the model's performance, we
use mapping ``funny'' \textrightarrow\ 1; ``normal'' \textrightarrow\ 0
such that standard metrics like accuracy can be computed.

\textbf{Influence Functions}.
Following \citet{han-tsvetkov-2020-fortifying}, we leverage influence
functions \citep{cook1982residuals,koh2017understanding} to identify the most influential training examples.
For each test example, we compute an influence score
of every training example.
Denoting an annotated training example as
$\mathbf{z}_{i}=\left(\mathbf{x}_{i}, y_{i}\right)$,
the optimal parameters obtained on the annotated data
$\{\mathbf{z}_{1}, \mathbf{z}_{2},...,\mathbf{z}_{n}\}$
as
$\hat{\theta} \stackrel{\text { def }}{=} \arg \min
_{\theta \in \Theta} \frac{1}{n} \sum_{i=1}^{n} L\left(\mathbf{z}_{i},
\theta\right)$,
where $\theta$ refers to model parameters,
$L\left(\mathbf{z}, \theta\right)$ refers to the loss,
and $n$ refers to the number of examples.
An intuitive way of computing the influence of
an example $\mathbf{z}_{i}$
is to inspect how $\hat{\theta}$ changes, when excluding
or including $\mathbf{z}_{i}$ into the training dataset; i.e.,
$\mathcal{I}(\mathbf{z}_{i})$ = $\hat{\theta}_{-\mathbf{z}_i}-\hat{\theta}$,
where
$\hat{\theta}_{-\mathbf{z}_i} \stackrel{\text { def }}{=}\arg\min_{\theta \in \Theta} \frac{1}{n-1} \sum_{ z \neq \mathbf{z}_{i}} L\left(\mathbf{z}_{i}, \theta\right)$.
But this is computationally expensive because it requires
retraining the model for every example $\mathbf{z}_{i}$.
\citet{cook1982residuals} propose
influence functions as an efficient
approximation by considering up-weighting
a training example $\mathbf{z}_{i}$
by a small value $\epsilon_i$.
Because we are more interested in the change of model loss/predictions, we
compute the influence of a training example $\mathbf{z}_{i}$
\emph{to the prediction of a test example} $\mathbf{z}_{\text {test}}$,
  following \citet{koh2017understanding}:
\begin{equation*}
\scriptsize
\label{inf_comp}
\begin{split}
    \mathcal{I}(\mathbf{z}_{i},\mathbf{z}_{\text {test }}) &\stackrel{\text {def}}{=}\frac{d L(\mathbf{z}_{\text {test }}, \hat{\theta})}{d \epsilon_{i}} \\
   &=-\nabla_{\theta} L(\mathbf{z}_{\text {test}}, \hat{\theta})^{\top}(\frac{1}{n} \sum_{j=1}^{n} \nabla_{\theta}^{2} L(\mathbf{z}_{j},\hat{\theta}))^{-1}\!\nabla_{\theta} L(\mathbf{z}_{i},\hat{\theta}).
\end{split}
\end{equation*}
For each test example $\mathbf{z}_{\text {test}}$, we compute an
influence score of each training example and then conduct
$z$-normalization.
We can then inspect whether or not the most influential
training examples contain offense and then correct or discard them.

\section{Datasets}
We leverage the human-annotated HaHackathon Dataset (\textbf{HHD})
of SemEval 2021 Task 7 \citep{meaney-etal-2021-semeval}.
An HHD example
consists of a text and two annotations: A binary
humor label and an offense score ranging from 1 to 5.
A humorous and not offensive example is
``Why do birds fly south in the Winter?  Because its too far to walk! ''
while a humorous but offensive example is
``Getting a girlfriend is a lot like getting a
car. The more money you have, the more options you have.''.


HHD contains humor texts covering eight categories including
Sexism, Body, Origin etc.
In this work, we limit our scope to gender-related humor,
and extract related examples with
surface form matching using the keywords in category ``Sexism''.
We call our constructed dataset Gender Humor Dataset (\textbf{GHD}).
We enforce two constraints when creating GHD:
\textbf{(i)} For gender-related texts and the ones not related to gender, we keep the same amount of humorous and non-humorous texts. This makes sure that a model cannot make predictions
through simple heuristics.
\textbf{(ii)}  We use 80\% examples for training and 20\% examples for
testing.
Appendix \figref{fig:offense_all} plots
the offense score distribution of
GHD training examples.  We observe a long-tail distribution,
where the maximum offense score is 4.7 and the average offense score
is 0.64.

\section{Experiment}
\seclabel{chap:exp}

We conduct a series of experiments to
answer the following research questions.
\textbf{RQ1}: How well does prompting
perform in humor recognition, compared
to finetuning when the amount of labeled data varies?
\textbf{RQ2}: Do the influential training examples
  correspond to humorous test examples contain offensive contents?
\textbf{RQ3}: What is the difference between the identified
offensive training examples in finetuning and prompting?

In our experiments, we use
\texttt{BERT-base-uncased} \citep{devlin2018bert} as our PLM.
We implement our models using PyTorch \citep{NEURIPS2019_9015} and HuggingFace \citep{wolf-etal-2020-transformers}.
Detailed experiment configurations are in Appendix
\secref{app:experimentconfig}.

\subsection{Comparing Finetuning and Prompting Humor Recognition Methods (RQ1)}
\seclabel{sec:finetune_prompt}
We firstly compare the performance of finetuning and prompting
of recognizing humor with HHD.
We conduct experiments using all of the training examples (``Full data'')
as well as randomly sampled few-shot data.
For few-shot experiments, we repeat each experiment five times
and report mean and standard deviation.
\figref{fig:two_methods} left shows the results.
It can be observed that prompting has an overall better performance
than finetuning in the few-shot scenarios; the performance difference
between these two methods decreases as more examples become available.
Lastly, prompting and finetuning perform similarly when all HHD
training examples are used.

\begin{figure}
\begin{minipage}{.5\linewidth}
\scriptsize
\begin{tabular}{ccccc}
    \toprule
     & \multicolumn{2}{c}{\textbf{Prompting}} &  \multicolumn{2}{c}{\textbf{Finetuning}}\\
    \cmidrule(lr){2-3} \cmidrule(lr){4-5}
     & \textbf{F1-score} & \textbf{Acc.} & \textbf{F1-score} & \textbf{Acc.} \\
    \midrule
    16-shot &  \textbf{74.20} (10.65) & \textbf{67.86} (5.57) & 67.32 (15.03) & 58.38 (4.07)\\
    32-shot &  \textbf{79.68} (\hphantom{0}1.15)  & \textbf{70.90} (0.96) & 75.66 (\hphantom{0}0.73)  & 62.28 (1.37) \\
    64-shot &  \textbf{80.36} (\hphantom{0}0.52)  &         73.21  (0.46) & 80.29 (\hphantom{0}2.25)  & \textbf{74.64} (3.66) \\
    128-shot &  \textbf{83.26} (\hphantom{0}0.83) & \textbf{78.08} (0.84) & 81.07 (\hphantom{0}2.88)  & 74.48 (4.32) \\
    \midrule
    Full data &  92.85 & 91.00 & \textbf{93.00} & \textbf{91.30} \\
    \bottomrule
\end{tabular}
\end{minipage}
\hspace{.5cm}
\begin{minipage}{.5\linewidth}
\centering
\includegraphics[width=.7\linewidth]{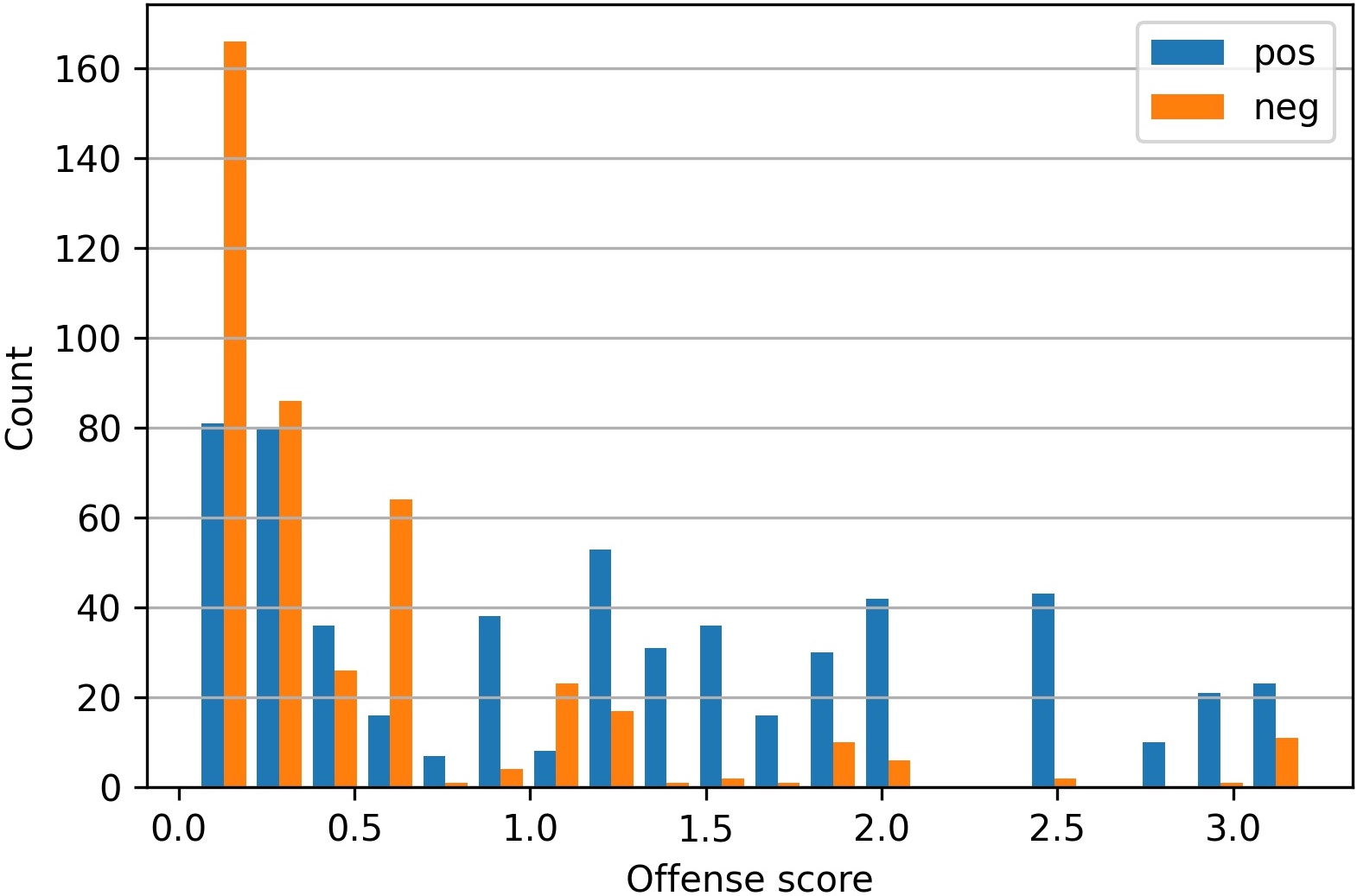}
\end{minipage}
\caption{Left: Humor recognition results (\%) of prompting and finetuning on
  HHD. We report mean and standard deviation (in parentheses) of task
  performance computed with five runs.  Right: offense score distribution
  of 30 most influential
  training samples.
}
\figlabel{fig:two_methods}
\end{figure}

\subsection{Offensive Influential Examples (RQ 2-3)}
In this section, we investigate if influence functions (\textsc{If}) can be used to
identify training examples that are sexism-offensive.
For a test example, we compute influence score
of every training example and then inspect
\emph{the most influential ones} to see if strong
offensive contents present.
Similar to section \secref{sec:finetune_prompt},
we compare the results of prompting and finetuning,
in full data and few-shot scenarios, but using GHD.
We adopt the same experiment configurations as \secref{sec:finetune_prompt}.

\subsubsection{Model Trained on Full Dataset}
\seclabel{sec:full_data}
Following \citet{han2020explaining}, we randomly sample 50 test examples
and identify their top 10, 30, and 50 most influential training examples.
We then report the averaged offensive score.
In addition, we use the average offensive score of all the GHD
training examples as a reference.

\begin{table}[t]
\centering\scriptsize
\begin{tabular}{@{}cccc@{}}
\toprule
                                                 & \textbf{TopK Influential}      & \textbf{Avg.Offen.(pos)}     & \textbf{Avg.Offen.(neg)} \\ \midrule
\multicolumn{1}{c}{\multirow{7}{*}{Full Data}} & \multicolumn{1}{c}{FT Top10}  & \multicolumn{1}{c}{1.39 ($\pm$0.96)} & \multicolumn{1}{c}{0.45 ($\pm$0.61)} \\
\multicolumn{1}{c}{}                           & \multicolumn{1}{c}{FT Top30}  & \multicolumn{1}{c}{1.23 ($\pm$0.93)} & \multicolumn{1}{c}{0.53 ($\pm$0.64)} \\
\multicolumn{1}{c}{}                           & \multicolumn{1}{c}{FT Top50}  & \multicolumn{1}{c}{1.16 ($\pm$0.89)} & \multicolumn{1}{c}{0.61 ($\pm$0.69)} \\ \cmidrule(l){2-4}
\multicolumn{1}{c}{}                           & \multicolumn{1}{c}{PT Top10}  & \multicolumn{1}{c}{1.02 ($\pm$0.93)} & \multicolumn{1}{c}{0.22 ($\pm$0.31)} \\
\multicolumn{1}{c}{}                           & \multicolumn{1}{c}{PT Top30}  & \multicolumn{1}{c}{0.95 ($\pm$0.82)} & \multicolumn{1}{c}{0.24 ($\pm$0.30)} \\
\multicolumn{1}{c}{}                           & \multicolumn{1}{c}{PT Top50}  & \multicolumn{1}{c}{0.94 ($\pm$0.85)} & \multicolumn{1}{c}{0.30 ($\pm$0.31)} \\ \cmidrule(l){2-4}
\multicolumn{1}{c}{}                           & \multicolumn{1}{c}{Reference} & \multicolumn{2}{c}{0.64}                                                   \\ \midrule
\multicolumn{1}{c}{\multirow{7}{*}{128-Shot}}  & \multicolumn{1}{c}{FT Top8}   & \multicolumn{1}{c}{1.24 ($\pm$1.24)} & \multicolumn{1}{c}{0.84 ($\pm$1.04)} \\
\multicolumn{1}{c}{}                           & \multicolumn{1}{c}{FT Top16}  & \multicolumn{1}{c}{1.33 ($\pm$1.27)} & \multicolumn{1}{c}{0.93 ($\pm$1.14)} \\
\multicolumn{1}{c}{}                           & \multicolumn{1}{c}{FT Top32}  & \multicolumn{1}{c}{1.33 ($\pm$1.27)} & \multicolumn{1}{c}{0.93 ($\pm$1.14)} \\ \cmidrule(l){2-4}
\multicolumn{1}{c}{}                           & \multicolumn{1}{c}{PT Top8}   & \multicolumn{1}{c}{1.53 ($\pm$1.26)} & \multicolumn{1}{c}{0.59 ($\pm$0.67)} \\
\multicolumn{1}{c}{}                           & \multicolumn{1}{c}{PT Top16}  & \multicolumn{1}{c}{1.49 ($\pm$1.25)} & \multicolumn{1}{c}{0.64 ($\pm$0.77)} \\
\multicolumn{1}{c}{}                           & \multicolumn{1}{c}{PT Top32}  & \multicolumn{1}{c}{1.30 ($\pm$1.21)} & \multicolumn{1}{c}{0.67 ($\pm$0.76)} \\ \cmidrule(l){2-4}
\multicolumn{1}{c}{}                           & \multicolumn{1}{c}{Reference} & \multicolumn{2}{c}{0.67}                                                   \\ \bottomrule
\end{tabular}
\caption{
  Averaged offense score of identified TopK influential training examples. As an example, the 
  top-3 ranked joke
  ``In many U.S. States offenders receive a harsher penalty for hitting a dog than they do for hitting a woman. That’s outrageous either way you’re slapping a bxxxh.'' is obviously offensive, with a score of 3.2.
  ``FT'': finetuning; ``PT'': prompting.
   Different amount of training examples are used: Full data and 128 training examples;
  ``Reference'' shows the averaged offensive score of training examples.
  ``pos'': humorous test examples; ``neg'': non-humorous test examples.
}
\tablabel{tab:identifiedinfluencescore}
\end{table}

\tabref{tab:identifiedinfluencescore}, ``Full Data'' rows,
shows that the most influential
training examples for humorous test examples (``pos'')
indeed have high offensive score.
For example, the average offense of the top
ten influential training examples is 1.39 for finetuning
and 1.02 for prompting, which are clearly higher than
the reference 0.64 of all the training examples.
On the other hand, for non-humorous test examples (``neg''), the
identified influential training examples have offense
scores smaller than reference: 0.45 for finetuning
and 0.22 for prompting.

In \figref{fig:two_methods} right, we show
the offense score distribution
of the top 30 most influential training examples for humorous
(``pos'') and non-humorous (``neg'') test examples.
For non-humorous examples,
most of the influential training examples
have low offense scores, ranging
from 0 to 1. However for humorous examples,
the offense scores spread out to different values,
achieving a maximum of 3.
Overall, these observations imply that
\emph{offensive content contributes to the presence of humor},
which clearly is an undesirable property given
the fact that there are non-offensive humorous
training examples.

Similar to \figref{fig:two_methods} left,
finetuning and prompting
perform similarly on identifying the influential training
examples in this full data scenario.

\subsubsection{Model Trained on Few-shot Data}

We also investigate the results of identifying influential training
examples in low resource
scenarios. \tabref{tab:identifiedinfluencescore} also shows results
when using 128 training examples.  We repeat each experiment three
times and then report the average.

In the 128-shot scenario, we observe that
both finetuning and prompting perform worse
in identifying correct influential training examples
than in the full data scenario.
This is more obvious for finetuning:
For non-humorous (``neg'') test examples,
the identified most influential
training examples have
high offensive scores (e.g, 0.84 of the top eight examples),
surpassing  the reference (0.67),
similar to the influential examples to
the humorous (``pos'') test examples.
This is undesirable because the model relies
on incorrect training examples for non-humorous test examples.
This reflects that
finetuning, which often requires numerous annotated data
to perform well,
cannot collaborate with influence functions
to accurately identify important training examples
in this low-resource scenario.

In contrast, combining prompting and influence functions gives more
accurate identification of influential and offensive training
examples: The most influential examples for humorous texts have
offensive scores higher than the reference and the most influential
examples for non-humorous texts have offensive scores lower than the
reference. Overall, \emph{prompting should be preferred in
recognizing humor and identifying influential offensive training examples
when the annotated data is limited.}

\section{Related Work}
Early work on humor recognition is based on
human-designed stylistic features,
e.g., alliteration chain,
semantic ambiguity, and semantic relatedness.
They have been shown to be very effective
\citep{mihalcea2005making,Attardo+2010}.
More recent works focus on
deep neural network models like CNN \citep{morales2017identifying,liu2018modeling,chen2017convolutional,chen2018humor,weller2019humor}.
State-of-the-art systems rely on
transfer learning with PLMs \citep{fan2020phonetics,xie2021humorhunter}.
E.g., DeepBlue \citep{song-etal-2021-deepblueai-semeval}
ensembles several finetuned RoBERTa \citep{liu2019roberta} and ALBERT \citep{Lan2020ALBERTAL} models,
achieving high performance of humor recognition
in SemEval \citep{meaney-etal-2021-semeval}.
In this work, we explore the effectiveness of
prompting \citep{GPT3paper,schick20cloze,liu2021pre,lisaprefix,Prompttuningpaper,xlmrprompting,Ptuningpaper,zhao-etal-2022-lmturk}, a
new transfer learning paradigm of utilizing PLMs, for humor recognition.
We show that
prompting performs better than (resp.\ comparably to)
finetuning in
low-resource (resp.\ high-resource)
scenarios.

Identifying offense in
computational humor is of high importance because
we want our NLP systems
to entertain users with no harm.
It is non-trivial to detect the
subtle relationship
between humor and offense \citep{meaney-etal-2021-semeval, hofmann2020gender, ruch2010sense};
we show that influence functions can be
leveraged to identify the offensive examples
and is more compatible with prompting than finetuning in
low-resource transfer learning scenarios.

\section{Conclusion}
This paper focuses on humor recognition systems
built upon PLMs and transfer learning.
We investigate the effectiveness of prompting for humor
recognition and show that it outperforms finetuning when annotations
are limited. By employing influence functions,
we characterize and show that models can rely on offense to recognize humor
during transfer -- which is highly undesirable
for real-world applications.
Future work may explore methods
of decreasing the offense,
and investigate offense related to
other aspects than gender/sexism
in humor recognition systems.

\section*{Acknowledgments}
This work was partially funded by the European Research Council (grant \#740516)
and the
German Federal Ministry of Education and Research (BMBF, grant \#01IS18036A).

\bibliography{bibpaper}
\appendix
\section{Appendix}
\subsection{Dataset Statistics}
In HHD training dataset, there are 4,932 humorous examples
and 3,068 non-humorous examples.
In HHD test dataset, there are 615 humorous
examples and 385 non-humorous examples.
For GHD dataset, we keep the same
proportion of humorous examples and non-humorous examples.
In GHD training
dataset, there are 766 humorous examples and 763 non-humorous examples.
In GHD test dataset, there are 190 humorous examples and 193 non-humorous
examples. We also show the offense score distribution of the examples
in GHD dataset in \figref{fig:offense_all}.
\begin{figure}[h!]
    \centering
    \includegraphics[width=.5\linewidth]{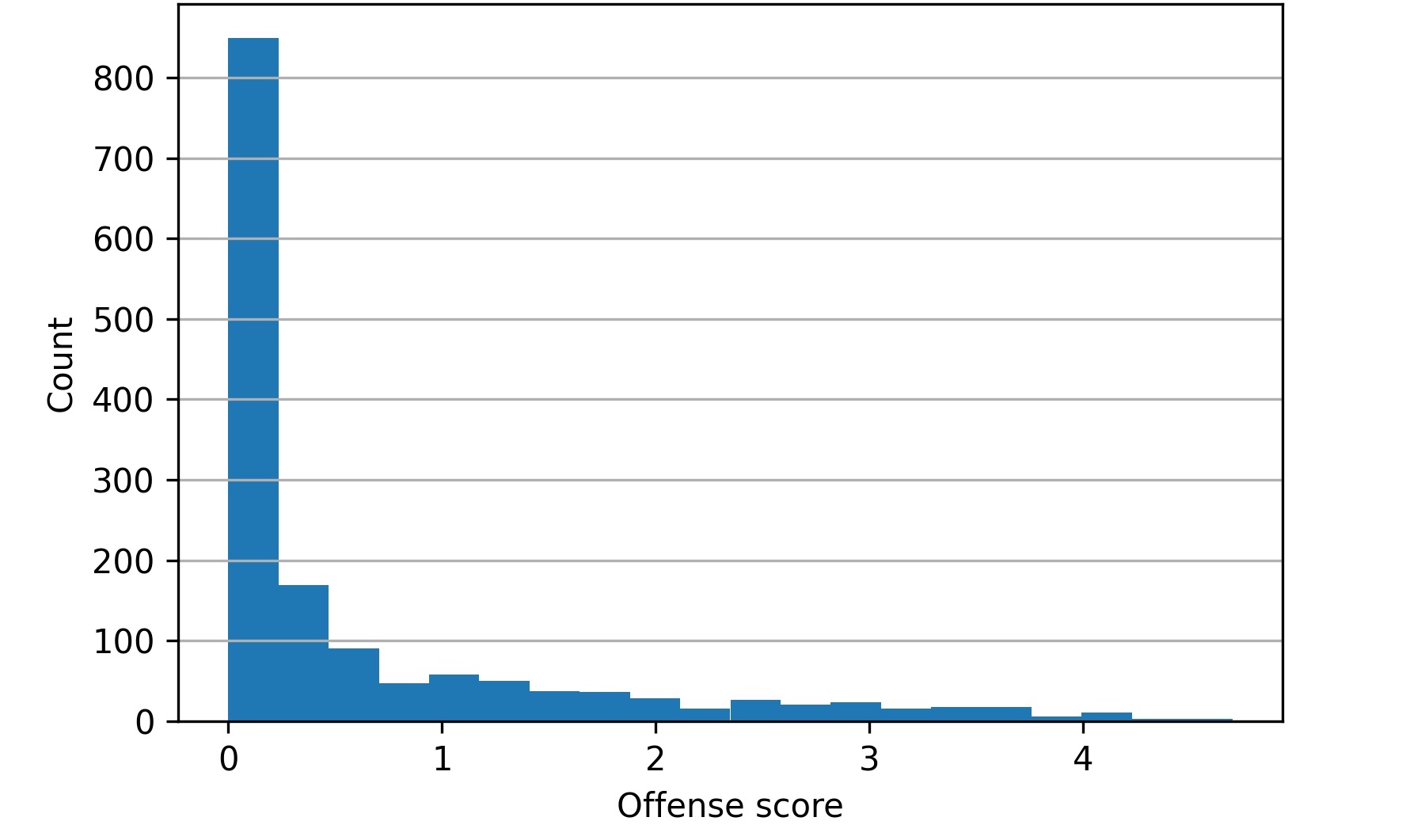}
    \caption{The offense score distribution of Gender Humor dataset}
    \figlabel{fig:offense_all}
\end{figure}

\subsection{Experiment Configurations}
\seclabel{app:experimentconfig} When we compute the influence function
results for finetuning and prompting, we use the same
experiment settings. We implement the influence function by adapting
the implementation of \citet{han2020explaining} using the LiSSA estimation
algorithm \citep{agarwal2017second}. Since deep neural network models
like BERT are not
convex, an additional damping term should be added to make sure that
the Hessian matrix is invertible and positive-definite. We choose the
damping term of $3\times10^{-3}$ as \citet{han-tsvetkov-2020-fortifying}.
The training batch size is 16, the
test batch size is 8, the learning rate is $5\times10^{-5}$, the
number of epochs is 3. To simplify the computation of the influence
function, we do not update the parameters in the embedding layer and
the first 8 Transformer layers, and only finetune the parameters in
remaining layers as \citet{han2020explaining}.
We load the parameters from the pretrained
BERT-base-uncased model and keep the max input text length of 128. All
experiments are conducted on a machine with Intel Core i7-6700K CPU,
Nvidia GeForce GTX 1080 GPU, and 16GB RAM.

\end{document}